\documentclass[letterpaper]{article}

\usepackage{aaai21}
\usepackage{times}
\usepackage{helvet}
\usepackage{courier}
\usepackage[hyphens]{url}
\usepackage{graphicx}
\usepackage{graphicx}
\usepackage{natbib}
\usepackage{caption}
\usepackage{amsfonts}
\usepackage{amssymb}
\usepackage{amsthm}
\usepackage{amsmath}
\usepackage{forest}
\PassOptionsToPackage{hyphens}{url}
\usepackage[backref=false,colorlinks=true,final,linkcolor=black,urlcolor=red!60!black,
citecolor=blue!65!black]{hyperref}
\usepackage{multirow}
\usepackage{xcolor}
\usepackage{xspace}
\newcommand{\comment}[1]{}
\renewcommand{\vec}[1]{\ensuremath{\boldsymbol{#1}}}
\DeclareMathOperator*{\argmax}{arg\,max}

\DeclareMathOperator*{\stable}{stable}

\DeclareMathOperator*{\fair}{fair}
\DeclareMathOperator*{\OH}{OH}
\DeclareMathOperator*{\HR}{HR}

\DeclareMathOperator*{\fairm}{\mathit{fair}}
\DeclareMathOperator*{\Split}{\mathit{split}}
\newcommand{\ud}{\triangleq}

\newcommand{\udr}{\stackrel{{\mbox{\tiny\ensuremath{\triangle}}}}{\Leftrightarrow}}
\newcommand{\tuple}[1]{\langle {#1}\rangle}

\newtheorem{theorem}{Theorem}[section]

\newtheorem{definition}[theorem]{Definition}

\newcommand{\silva}{Silva\xspace}
\newcommand{\meta}{Meta-Silvae\xspace}
\newcommand{\FATT}{FATT\xspace}

\title{Fair Training of Decision Tree Classifiers}
\author{
 Francesco Ranzato\textsuperscript{\rm 1}, 
 Caterina Urban\textsuperscript{\rm 2}, 
 Marco Zanella\textsuperscript{\rm 1} \\
 } 
 
 \affiliations {
 \textsuperscript{\rm 1} Dipartimento di Matematica, University of Padova, Italy\\ \{ranzato, mzanella\}@math.unipd.it \\
 \textsuperscript{\rm 2} INRIA and \'{E}cole Normale Sup\'{e}rieure $|$ Universit\'{e} PSL, France \\
 caterina.urban@inria.fr
}

\begin{document}

\maketitle

%-----------------------------------------------------------------------
\begin{abstract}
We study the problem of formally verifying individual fairness of decision tree ensembles, as well as training tree models which maximize both accuracy and individual fairness. In our approach, fairness verification and fairness-aware training both rely on a notion of stability of a classification model, which is a  variant of standard robustness under input perturbations used in adversarial machine learning. Our verification and training methods leverage abstract interpretation, a well established technique for static program analysis which is able to automatically infer assertions about stability properties of decision trees. By relying on a tool 
for adversarial training of decision trees, our fairness-aware learning method has been implemented 
 and experimentally evaluated on the reference datasets used to assess fairness properties. The experimental results show that our approach is able to train tree models exhibiting a high degree of 
individual fairness w.r.t.\ the natural state-of-the-art CART trees and
random forests. Moreover, as a by-product, these fair decision trees turn out to be significantly compact, thus enhancing the interpretability of their fairness properties.
% \medskip
% \noindent
% \textit{Content Areas:} to do
\end{abstract}

%-----------------------------------------------------------------------
\section{Introduction}
\label{sec:introduction}

The widespread adoption of data-driven automated decision-making software with far-reaching societal impact, e.g., for credit scoring \cite{Khandani}, recidivism prediction \cite{Chouldechova}, or hiring tasks~\cite{Schumann}, raises concerns 
on the fairness properties of these tools \cite{Barocas}.
Several fairness verification and bias mitigation approaches for machine learning (ML) systems have been proposed in recent years, e.g.~\cite{aghaei19,Grari,roh20,ruoss2020learning,Urban,yurochkin20,Zafar} among the others. However, most works focus on neural networks \cite{roh20,ruoss2020learning,Urban,yurochkin20} or on group-based notions of fairness \cite{Grari,Zafar}, e.g., demographic parity \cite{dwork2012fairness} or equalized odds \cite{Hardt}. These notions of group-based fairness require some form of statistical parity (e.g. between positive outcomes) for members of different protected groups (e.g.\ gender or race). On the other hand, they do not provide guarantees for individuals or other subgroups. By contrast, in this paper we focus on \emph{individual fairness}~\cite{dwork2012fairness}, intuitively meaning that similar individuals in the population receive similar outcomes, and on decision tree ensembles \cite{breiman-random-forests,friedman2001greedy}, which  are commonly used for tabular datasets since they are easily interpretable
ML models with high accuracy rates.  

\paragraph{Contributions.}

We propose an approach for verifying individual fairness of decision tree ensembles, 
%such as Random Forests (RFs) \cite{breiman-random-forests} and Gradient Boosted Decision Trees (GBDTs) \cite{friedman2001greedy}, 
as well as training tree models which maximize both accuracy and fairness.
The approach is based on \emph{abstract interpretation} \cite{CC77,rival-yi}, a well known static program analysis technique, and
builds upon a framework for training robust decision tree ensembles called \meta \cite{gat}, which in turn leverages a verification tool for robustness properties of decision trees~\cite{DBLP:conf/aaai/RanzatoZ20}. Our approach is fully parametric on a given underlying abstract domain representing input space regions containing similar individuals. We instantiate it with a product of two abstract domains: (a) the well-known 
abstract domain of hyper-rectangles (or boxes)~\cite{CC77}, that represents exactly the standard notion of similarity between individuals 
based on the $\ell_\infty$ 
distance metric, and does not lose precision 
for the univariate split decision rules of type $x_i\leq k$; and (b) a specific 
relational abstract domain which accounts for one-hot encoded categorical features. 

Our Fairness-Aware Tree Training method, called \FATT, is designed as an extension of \meta \cite{gat}, 
a learning methodology for ensembles of decision trees based on a genetic algorithm which is able to train a decision tree for maximizing both its accuracy and its robustness to adversarial perturbations.
We demonstrate the effectiveness of \FATT in training accurate and fair models on the standard datasets used in the literature on fairness. Overall, the experimental results show that our fair-trained models are on average between $35\%$ and $45\%$ more fair than naturally trained decision tree ensembles at an average cost of $-3.6\%$ of accuracy. Moreover, it turns out that our tree models are orders of magnitude more compact and thus more interpretable. Finally, we show how our models can be used as ``hints'' for setting the size and shape hyper-parameters (i.e., 
maximum depth and minimum number of samples per leaf) when training standard decision tree models. As a result, this hint-based strategy is capable to output models that are about $20\%$ more fair and just about $1\%$ less accurate than standard models.
%thus halfway in performance between standard models and our models trained with \meta).

\paragraph{Related Work.}
The most related work to ours is by Aghaei et al.~\cite{aghaei19},
Raff et al. \cite{Raff} and Ruoss et al. \cite{ruoss2020learning}.

By relying on the mixed-integer optimization learning approach by Bertsimas and Dunn~\cite{Bertsimas17}, Aghaei et al.~\cite{aghaei19}  put forward a
framework for training fair decision trees for classification and regression.
%, where the trade-off between accuracy and fairness can be tuned as a parameter.
The experimental evaluation shows that this approach mitigates unfairness as modeled by their notions of 
disparate impact and disparate treatment at the cost of a  significantly
higher training computational cost. 
Their notion of disparate treatment is distance-based and thus akin to individual fairness with respect to the nearest individuals \emph{in a given dataset} (e.g., the $k$-nearest individuals). In contrast, we consider individual fairness with respect to the nearest individuals \emph{in the input space} (thus, also individuals that are not necessarily part of a given dataset).

Raff et al. \cite{Raff} propose a regularization-based approach for training fair decision trees as well as fair random forests.  They consider both group fairness as well as individual fairness with respect to the $k$-nearest individuals in a given dataset, similarly to Aghaei et al.~\cite{aghaei19}.
In their experiments they use a subset of the datasets that we consider in our evaluation (i.e., the Adult, German, and Health datasets). 
Our fair models have higher accuracy than theirs (i.e., between $2\%$ and $5.5\%$) for all but one of these datasets (i.e., the Health dataset). Interestingly, their models (in particular those with worse accuracy than ours) often have accuracy on par with a constant classifier due to the highly unbalanced label distribution of the datasets (cf.\ Table~\ref{tab:datasets}). 

Finally, Ruoss et al.~\cite{ruoss2020learning} have proposed an approach for learning individually fair data representations and training neural networks (rather than decision tree ensembles as we do) that satisfy individual fairness with respect to a given similarity notion. We use the same notions of similarity in our experiments (cf. Section~\ref{subsec:similarity}).  %However, their approach is designed for neural networks rather than decision tree ensembles. 

%-----------------------------------------------------------------------
\section{Background}
\label{sec:background}
Given an input space $X \subseteq \mathbb{R}^d$ of numerical vectors and a finite set of labels $\mathcal{L} = \{y_1, \ldots, y_m\}$, a classifier is a function $C\colon X \rightarrow \wp_+(\mathcal{L})$, where $\wp_+(\mathcal{L})$ is the set of nonempty subsets of $\mathcal{L}$, which associates at least one label to every input in $X$. A training algorithm takes as input a dataset $D \subseteq X \times \mathcal{L}$  and outputs a classifier $C\colon X \rightarrow \wp_+(\mathcal{L})$ which optimizes some objective function, such as the Gini index or the information gain for decision trees. 

Categorical features can be converted into numerical ones through one-hot encoding, where a single feature with $k$ possible distinct categories $\{c_1,...,c_k\}$ 
is replaced by $k$ new binary features with values in $\{0, 1\}$.
Then, each value $c_j$ of the original categorical feature is represented by a bit-value assignment to the new $k$ binary features in which 
the $j$-th feature is set to $1$ (and the remaining $k-1$ binary features are set to $0$). 
%For instance, an attribute $\textit{color} \in \{\textit{red}, \textit{green}, \textit{blue}\}$ is replaced by $\textit{color}_{\textit{red}}, \textit{color}_{\textit{green}}, \textit{color}_{\textit{blue}} \in \{0, 1\}$
%and $\textit{color}=\textit{red}$ 
%is represented by $\textit{color}_{\textit{red}} = 1$, $\textit{color}_{\textit{green}} = 0$, $\textit{color}_{\textit{blue}} = 0$.

Classifiers can be evaluated and compared through several metrics. Accuracy on a test set is a basic metric: given a ground truth 
test set $T \subseteq X \times \mathcal{L}$, the accuracy of $C$ on $T$ is
$\mathit{acc}_{T}(C) \ud |\{(\vec{x}, y) \in T ~|~ C(\vec{x}) = \{y\}\}|/|T|$.
%One typically aims at training classifiers having a nearly perfect accuracy on suitably crafted test sets. 
However, according to a growing belief~\cite{cacm18}, accuracy is not enough in machine learning, since robustness to adversarial inputs 
of a ML classifier may significantly affect its safety and generalization properties~\cite{carlini,cacm18}.
Given an input perturbation modeled by a function  $P\colon X \rightarrow \wp(X)$, a classifier $C: X \rightarrow \wp_+(\mathcal{L})$ is 
\emph{stable} \cite{DBLP:conf/aaai/RanzatoZ20} on the perturbation $P(\vec{x})$ of $\vec{x} \in X$ when $C$ consistently assigns the same label(s) to every attack ranging in $P(\vec{x})$, i.e., 
\[
\stable(C, \vec{x}, P) \udr \forall \vec{x}' \in P(\vec{x})\colon C(\vec{x}') = C(\vec{x}).
\]
 When the sample $\vec{x}\in X$ 
has a ground truth label $y_{\vec{x}}\in \mathcal{L}$, robustness of $C$ on $\vec{x}$ 
boils down to 
stability $\stable(C, \vec{x}, P)$ together with correct classification
$C(\vec{x}) = \{y_{\vec{x}}\}$.

We consider standard classification decision trees commonly referred to as CARTs  (Classification And Regression Trees)~\cite{BreimanFOS84}. A decision tree $t\colon X \rightarrow \wp_+(\mathcal{L})$ is defined inductively. A base tree $t$ is a single leaf $\lambda$ storing a frequency distribution of labels for the samples of the training dataset, hence $\lambda \in [0,1]^{|\mathcal{L}|}$, or, equivalently, $\lambda\colon \mathcal{L} \rightarrow [0,1]$. Some algorithmic rule converts this frequency distribution into a set of labels, typically  as $\argmax_{y \in \mathcal{L}}\lambda(y)$.
A composite tree $t$ is $\gamma(\Split, t_l, t_r)$, where $\Split\colon X \rightarrow \{\mathbf{tt}, \mathbf{ff}\}$ is a Boolean split criterion for the internal parent node of its left and right subtrees $t_l$ and $t_r$; thus, for all $\vec{x} \in X$, $t(\vec{x}) \ud \textbf{if}~\Split(\vec{x})~\textbf{then}~t_l(\vec{x})~\textbf{else}~t_r(\vec{x})$.  Although split rules can be of any type, most decision trees employ univariate hard splits of type $\Split(\vec{x}) \ud \vec{x}_i \leq k$ for some feature $i \in [1, d]$ and threshold $k \in \mathbb{R}$.

Tree ensembles, also known as forests, are sets of decision trees which together contribute to formulate a unique classification output. Training algorithms as well as methods for computing the final output label(s) vary among different tree ensemble models. Random forests (RFs) \cite{breiman-random-forests} are a major instance of tree ensemble where each tree of the ensemble is  trained  independently from the other trees on a random subset of the features. Gradient boosted decision trees (GBDT) \cite{friedman2001greedy} represent a different training algorithm where an ensemble of trees is incrementally build by training each new tree on the basis of the data samples which are mis-classified by the previous trees. For RFs, the final classification output is typically obtained through a voting mechanism (e.g., majority voting), while GBDTs are usually trained for binary classification problems and use some binary reduction scheme, such as one-vs-all or one-vs-one, for multi-class classification.

%-----------------------------------------------------------------------
\section{Individual Fairness}
\label{sec:individual-fairness}
Dwork et al. \cite{dwork2012fairness} define \emph{individual fairness} as ``the principle that two individuals who are similar with respect to a particular task should be classified similarly''. 
They formalize this notion as a Lipschitz condition of the classifier, which requires that any two individuals $\vec{x}, \vec{y} \in X$ whose 
distance is $\delta(\vec{x}, \vec{y}) \in [0, 1]$ map to distributions $D_{\vec{x}}$ and $D_{\vec{y}}$, respectively, such that the statistical distance between $D_{\vec{x}}$ and $D_{\vec{y}}$ is at most $\delta(\vec{x}, \vec{y})$. The intuition is that 
the output distributions for   $\vec{x}$ and 
$\vec{y}$ are indistinguishable up to their distance 
$\delta(\vec{x}, \vec{y})$.
The distance metric
$\delta\colon X \times X \rightarrow \mathbb{R}_{\geq 0}$ is problem specific and 
satisfies the basic axioms $\delta(\vec{x}, \vec{y}) = \delta(\vec{y}, \vec{x})$ and $\delta(\vec{x}, \vec{x}) = 0$.
%for all $\vec{x}, \vec{x}, \vec{y} \in X$. 

By following Dwork et al's standard definition~\cite{dwork2012fairness}, we consider a classifier $C\colon X \rightarrow \wp_+(\mathcal{L})$ to be fair 
when $C$ outputs the same set of labels for every pair of individuals $\vec{x}, \vec{y} \in X$ which satisfy a similarity relation $S \subseteq X \times X$. Thus,  $S$ can be derived from a distance $\delta$ as $(\vec{x}, \vec{y}) \in S \udr \delta(\vec{x}, \vec{y}) \leq \epsilon$, where $\epsilon \in \mathbb{R}$ is a threshold of similarity. In order to estimate a fairness metric for a classifier $C$, we count how often $C$ is fair on sets of similar individuals ranging into a test set $T \subseteq X \times \mathcal{L}$:
\begin{equation}\label{eq:fairness}
\textstyle{\fairm_{T}(C)} \ud 
\displaystyle
\frac{|\{(\vec{x}, y) \in T ~|~ \fair(C, \vec{x}, S)\}|}{|T|}
\end{equation}
where 
$\fair(C, \vec{x}, S)$ is defined as follows:

\begin{definition}[\textbf{Individual Fairness}]\rm
\label{def:individual-fairness}
 A classifier $C\colon X \rightarrow \wp_+(\mathcal{L})$ is \emph{fair} on an input sample $\vec{x} \in X$ with respect to a similarity relation $S \subseteq X\times X$, denoted by $\fair(C, \vec{x}, S)$, when $\forall \vec{x}' \in X\colon (\vec{x}, \vec{x}') \in S \Rightarrow C(\vec{x}') = C(\vec{x})$. \qed
\end{definition}

Hence, fairness for a similarity relation $S$ 
boils down to stability
on the perturbation $P_S(\vec{x}) \ud 
\{\vec{x}' \in X ~|~ (\vec{x}, \vec{x}') \in S\}$, 
namely, for all $\vec{x} \in X$, 
\begin{equation}
\label{individual-fairness-stability}
\fair(C, \vec{x}, S) \;\Leftrightarrow\;  \stable(C, \vec{x}, P_S)
\end{equation}

%\begin{proof}
% We first define $P_S(\vec{x}) = \{\vec{x'} \in X ~|~ (\vec{x}, \vec{x'}) \in S\}$. Clearly $(\vec{x}, \vec{x'}) \in S$ if and only if $\vec{x'} \in P_S(\vec{x})$. We rewrite $\fair(\vec{x}, C, S)$ as $\forall \vec{x'} \in X\colon (\vec{x}, \vec{x'}) \in S \Rightarrow C(\vec{x'}) = C(\vec{x})$, which is equivalent to $\forall \vec{x'} \in X\colon \vec{x'} \in P_S(\vec{x}) \Rightarrow C(\vec{x'}) = C(\vec{x})$ and $\forall \vec{x'} \in P_S(\vec{x})\colon C(\vec{x'}) = C(\vec{x})$. The latter is precisely the definition of $\stable(C, \vec{x}, P_S(\vec{x}))$. After unfolding $P_S(\vec{x})$ we get $\stable(C, \vec{x}, \{\vec{x'} \in X ~|~ (\vec{x}, \vec{x'}) \in S\})$.
%\end{proof}

Let us remark that fairness is orthogonal to accuracy since it 
does not depend on the correctness of the label assigned by the classifier, so that 
that training algorithms that maximize accuracy-based metrics do not necessarily achieve fair models. Thus, this is also the case of a natural learning algorithm for CART trees and RFs, that locally optimizes split criteria by measuring entropy or Gini impurity, which are both indicators of the correct classification of training data. 
% \todo{What are we trying to say with this?}

It is also worth observing that fairness is monotonic with respect to the similarity relation, meaning that 
\begin{equation}
\label{individual-fairness-subset}
\fair(C, \vec{x}, S) \wedge  S' \subseteq S \;\Rightarrow\; \fair(C, \vec{x}, S')
\end{equation}

%since a classifier which is stable on a region is also stable on every subregion:
%
%\begin{corollary}[Individual Fairness on Subsets]
%\label{th:individual-fairness-subset}
% If $\fair(C, \vec{x}, S)$ holds, then for every similarity (sub-)relation,  also holds.
%\end{corollary}
%
%\begin{proof}
%By \eqref{individual-fairness-stability}, $\fair(C, \vec{x}, S)$ is equivalent to $\stable(C, \vec{x}, \{\vec{x'} \in X ~|~ (\vec{x}, \vec{x'}) \in S\})$, which implies $\stable(C, \vec{x}, \{\vec{x'} \in X ~|~ (\vec{x}, \vec{x'}) \in S'\})$ for any $S' \subseteq S$, hence $\fair(C, \vec{x}, S')$.
%\end{proof}

\noindent
We will exploit this monotonicity property, since 
this implies that, on one hand,  
fair classification is preserved for smaller similarity relations and, on the other hand,
fairness verification and fair training is more challenging for  
larger similarity relations.  
%{\color{red}\bf +++TODO: WHERE AND HOW PRECISELY?+++}

%-----------------------------------------------------------------------
\section{Verifying Fairness}
\label{sec:fairness-verification}
As individual fairness is equivalent to stability, 
individual fairness of decision trees can be verified by \silva \cite{DBLP:conf/aaai/RanzatoZ20}, an abstract interpretation-based algorithm for checking 
stability properties of decision tree ensembles. 
\subsection{Verification by \silva}
\silva performs a static analysis of an ensemble of decision trees in a so-called abstract domain $A$ that approximates properties of real vectors, meaning that each abstract value $a\in A$ represents a set of real vectors $\gamma(a)\in\wp (\mathbb{R}^d)$.
\silva approximates an 
input region $P(\vec{x}) \in \wp(\mathbb{R}^d)$ for an input vector $\vec{x} \in \mathbb{R}^d$ by an abstract value $a\in A$ such that $P(\vec{x})\subseteq \gamma(a)$ and for each decision tree $t$, it computes an over-approximation of the set of leaves of $t$ that can be 
reached from some vector in 
$\gamma(a)$. This is computed by 
collecting the constraints of split nodes 
for each root-leaf path, so that each leaf $\lambda$ of $t$ stores
the minimum set of constraints $C_\lambda$ which makes $\lambda$ reachable from the root of $t$.
It is then checked if this set of constraints $C_\lambda$ can be satisfied by 
the input abstract value $a\in A$: this check is 
denoted by $a\models^? C_\lambda$ and its \emph{soundness} requirement means 
that if some input sample $\vec{z}\in \gamma(a)$ may reach the leaf $\lambda$ then 
$a\models C_\lambda$ must necessarily hold. 
When  $a\models C_\lambda$ holds the leaf 
$\lambda$ is marked as
reachable from $a$. For example, if $C_\lambda =\{ x_1\leq 2, 
\neg(x_1\leq -1), 
x_2 \leq -1\}$ then an abstract value such as 
$\tuple{x_1 \leq 0,x_2 \leq 0}$ satisfies $C_\lambda$ while a relational abstract value such as
$x_1 +x_2 =4$ does not. 
This over-approximation of the set of leaves of $t$ reachable from $a$ allows us to compute a set of labels, denoted by $t^A(a)\in \wp_+(\mathcal{L})$ which
is an over-approximation of the set of labels assigned by $t$ to all the input 
vectors ranging in $\gamma(a)$, i.e., 
$\cup_{\vec{z}\in \gamma(a)} t(\vec{z})\subseteq t^A(a)$ holds. Thus, if $P(\vec{x})\subseteq \gamma(a)$ and
$t^A(a) =t(\vec{x})$ then 
$t$ is stable on $P(\vec{x})$.  

For standard classification trees with hard univariate splits of type  $x_i \leq k$, 
we will use the well-known hyper-rectangle abstract domain  $\HR$ 
whose abstract values for vectors $\vec{x}\in \mathbb{R}^d$ are of type
\[h=\tuple{\vec{x}_i\in[l_1, u_1], \dots, \vec{x}_d\in [l_d, u_d]}\in \textstyle\HR_d\] 
where lower and upper bounds $l,u\in 
\mathbb{R} \cup \{-\infty,+\infty\}$ with $l\leq u$ (more on this abstract domain can be found in \cite{rival-yi}). Thus, $\gamma(h) = \{\vec{x}\in \mathbb{R}^d \mid \forall i.\, l_i \leq \vec{x}_i \leq u_i\}$.
The hyper-rectangle abstract domain   
guarantees that for each leaf 
constraint $C_\lambda$ and $h\in \HR$, the
check $h\models^? C_\lambda$ is (sound and) \emph{complete}, meaning that 
$h\models C_\lambda$ holds iff there exists some input sample in $\gamma(h)$ reaching $\lambda$. This completeness property  therefore entails that the set of labels $t^{\HR}(h)$ computed by 
this analysis coincides exactly with 
the set of classification labels computed by $t$ for all the samples in $\gamma(h)$, 
so that for the $\ell_\infty$-based perturbation such that 
$P_\infty(\vec{x})=\gamma(h)$ then it turns out that
$t$ is stable on $P_\infty(\vec{x})$ iff
$t^{\HR}(h) = t(\vec{x})$ holds. 

In order to analyse a forest $F$ of trees, \silva reduces the whole forest to a single tree $t_F$, by stacking every tree $t\in F$ on top of each other, i.e., 
each leaf becomes the root of the next tree in $F$, where 
the ordering of this unfolding operation does not matter. Then, each leaf $\lambda$ of this huge single tree $t_F$ collects all the constraints of the leaves in the path from the root of $t_F$ to $\lambda$. Since this stacked tree $t_F$ suffers from a combinatorial explosion of the number of leaves, \silva deploys a number of optimisation strategies for its analysis. Basically, \silva exploits a best-first search algorithm to look for a pair of input samples in $\gamma(a)$ which are differently labeled, hence showing instability. If one such instability counterexample can be found then instability is proved and the analysis terminates, otherwise stability is proved. Also, \silva allows to set a safe timeout which, when met, stops the analysis and outputs the current sound 
over-approximation of labels. 
% \todo{What does this counterexample look like? This is linked to the issue I have with the previous sentence.} Otherwise, the classifier is certified to be stable on $P(\vec{x})$.

%An example of abstract domain is that of hyper-rectangles $\mathcal{H}$ \cite{CC77}, whose elements are $d$-dimensional vectors of intervals $[l, u]$, where $l \in \mathbb{R} \cup \{-\infty\}$, $r \in \mathbb{R} \cup \{+\infty\}$, and $l \leq r$. An hyper-rectangle $([l_1, u_1], \dots, [l_d, u_d]) \in \mathcal{H}$ represents all real vectors $\vec{x} \in \mathbb{R}^d$ such that, for all $i \in [1, d]$, $l_i \leq \vec{x}_i \leq u_i$, e.g., such that lower- and upper-bounds of their components are correctly approximated by the hyper-rectangle.

\subsection{Verification with One-Hot Enconding}
As described above, the soundness of \silva guarantees that no true reachable leaf is missed by this static analysis. Moreover, when the input region $P(\vec{x})$ is defined by the $\ell_\infty$ norm and the static analysis is performed using the abstract domain of hyper-rectangles $\HR$, \silva is also complete, meaning that no false positive (i.e., a false reachable leaf) can occur. However, that this is not true anymore when dealing with classification problems involving some categorical features.

%\begin{figure}[ht]
 \begin{center}
 \begin{forest} 
  [$\textit{color}_{\textit{white}} \leq 0.5$
    [$\{\ell_1\}$, name = n1]
    [$\{\ell_2\}$]
  ]
   \node at (n1) [left=6ex,above=1ex]{{\normalsize $t_1$}};
 \end{forest}
 \quad
 \begin{forest} 
  [$\textit{color}_{\textit{black}} \leq 0.5$
    [$\{\ell_2\}$]
    [$\{\ell_1\}$, name = n2]
  ]
  \node at (n2) [right=6ex,above=1ex]{{\normalsize $t_2$}};
 \end{forest}
 \end{center}
% \caption{A random forest with two trees.} 
%\label{fig:random-forest}
%\end{figure}

\noindent
The diagram above depicts a toy forest $F$ consisting of two trees $t_1$ and $t_2$, where left/right branches are followed when the split condition is false/true. Here, a categorical feature $\textit{color} \in \{\textit{white}, \textit{black}\}$ is one-hot encoded by $\textit{color}_{\textit{white}}, \textit{color}_{\textit{black}} \in \{0, 1\}$. Since colors are mutually exclusive, every white individual in the input space, i.e.\ 
$\textit{color}_{\textit{white}}=1,\textit{color}_{\textit{black}}=0$, 
will be labeled as $\ell_1$ by both trees. However, by running the stability analysis on the hyper-rectangle $\langle\textit{color}_{\textit{white}}\in [0,1], \textit{color}_{\textit{black}}\in [0, 1]\rangle$, \silva would mark the classifier as unstable because there exists a sample in $[0,1]^2$ whose output is $\{\ell_1, \ell_2\} \neq \{\ell_1\}=F(\textit{color}_{\textit{white}}=1,\textit{color}_{\textit{black}}=0)$. This is due to the point $(0, 0) \in [0, 1]^2$ which is a feasible input sample for the analysis, although it does not represent any actual individual in the input space. In fact, 
$t_1(0,0)=\{\ell_2\}$ and
$t_2(0,0)=\{\ell_1\}$, 
so that by a majority voting $F(0,0)=\{\ell_1, \ell_2\}$, thus making $F$ unstable (i.e.,
unfair) on 
$(1,0)$ (i.e., on white individuals).

To overcome this issue, we instantiate \silva to an abstract domain which is designed as
a reduced product (more details on reduced products can be found in \cite{rival-yi})  with  a relational abstract domain 
keeping track of the relationships among the multiple 
binary features introduced by one-hot encoding a categorical feature. 
More formally, this relational domain maintains the following 
two additional constraints on the $k$ features $x^c_1,...,x^c_k$ introduced by one-hot encoding a categorical variable $x^c$ with $k$ distinct values: 
\begin{enumerate}
\item %[{\rm (1)\/}] 
the possible values for each $x^c_i$ are restricted to $\{0,1\}$; 
\item %[{\rm (2)\/}] 
the sum of all  $x^c_i$ must satisfy $\sum_{i=1}^k x^c_i =1$.
\end{enumerate}
Hence, these conditions guarantee that any abstract value for $x^c_1,...,x^c_k$ represents
precisely one possible category for $x^c$. This abstract domain 
for a categorical variable $x$ with $k$ distinct values is denoted by $\OH_k(x)$.
In the example above, any hyper-rectangle $\langle\textit{color}_{\textit{white}}\in [0,1], \textit{color}_{\textit{black}}\in [0, 1]\rangle$ is reduced by $\OH_2(\textit{color})$, so 
that just two different values 
$\langle\textit{color}_{\textit{white}}= 0, \textit{color}_{\textit{black}}=1\rangle$
and  $\langle\textit{color}_{\textit{white}}= 1, \textit{color}_{\textit{black}}=0\rangle$
are allowed. 

Summing up, the generic abstract value of the reduced hyper-rectangle domain 
computed by the analyzer \silva for 
data vectors consisting of $d$ numerical variables  $x^j\in \mathbb{R}$ 
and $m$ categorical variables $c^j$ with $k_j\in \mathbb{N}$ distinct values is:
\begin{equation*}
\textstyle
\tuple{x^j\in[l_j, u_j]}_{j=1}^d \times 
\tuple{c^j_{i}\in \{0,1\} \mid \sum_{i=1}^{k_j}  c^j_{i}=1}_{j=1}^m
\end{equation*}
where $l_j,u_j\in 
\mathbb{R} \cup \{-\infty,+\infty\}$ and $l_j\leq u_j$.

%-----------------------------------------------------------------------
\section{\FATT: Fairness-Aware Training of Trees}
\label{sec:fairness-training} 
Several algorithms for training robust decision trees and ensembles have been put forward \cite{Andriushchenko19, calzavara2019B, calzavara20, ChenZBH19, kantchelian, gat}. These algorithms encode the robustness of a tree classifier as a loss function which is minimized either by 
either exact methods such as MILP or by suboptimal heuristics such as genetic algorithms. 

The robust learning algorithm of \cite{gat}, called \meta, 
aims at maximizing
a tunable weighted linear combination of accuracy and stability metrics. \meta relies on a genetic algorithm for evolving a population of trees which are ranked by their accuracy and stability, where tree stability is computed
by resorting to the verifier \silva~\cite{DBLP:conf/aaai/RanzatoZ20}. At the end of this genetic evolution, \meta returns the best tree(s).  It turns out that \meta typically outputs compact models which are easily interpretable and often achieve accurate and stable models already with a single decision tree rather than a forest. By exploiting the equivalence~\eqref{individual-fairness-stability}
between individual fairness and stability and
the instantiation of the verifier \silva to the  product abstract domain tailored for one-hot encoding, we use \meta as a learning algorithm for decision trees, called \FATT, that 
enhances their individual fairness.  

%as a black box, all of the adaption implemented in \silva and described in Sec.  directly carry over to \meta. Only modification to meta silvae involved its command line interface, for which we introduced an additional option to configure which features are part of the same one-hot encoding; this information is then directly passed to silva.
% \todo{Mh. This part needs some extending.}

While standard learning algorithms for tree ensembles
require tuning some hyper-parameters, such as maximum depth of trees, minimum amount of information on leaves and maximum number of trees, \meta is able to infer them 
automatically, so that the traditional tuning process is not needed. Instead, some standard parameters are required by the underlying 
genetic algorithm, notably, the size of the evolving population, the maximum number of evolving iterations, the crossover and mutation functions 
\cite{holland1984genetic,Srinivas}. Moreover, we need to specify 
the objective function of \FATT that, for learning fair decision trees, 
is given by a weighted sum of the accuracy and individual 
fairness scores over the training set. 
It is worth remarking that, given an objective function, the genetic algorithm of \FATT 
converges to an optimal (or suboptimal) solution regardless of the chosen parameters, which
just affect the rate of convergence and therefore 
should be chosen for tuning its speed. 

Crossover and mutation functions are two main distinctive features of the genetic algorithm of \meta. 
The crossover function of \meta  combines two parent trees $t_1$ and $t_2$ by randomly substituting a subtree of $t_1$ with a subtree of $t_2$. Also, \meta supports
two types of mutation strategies: grow-only, which only allows trees to grow, and grow-and-prune, 
which also allows pruning the mutated trees. Finally, let us point out that \meta allows to set the basic
parameters used by generic algorithms: population size, selection function,
number of iterations.
In our instantiation of \meta to fair learning: the population size 
is kept fixed to $32$, as the experimental evaluation showed that this provides an effective 
balance between achieved fairness and training time; 
the standard roulette wheel algorithm is employed as selection function; the 
number of iterations is typically dataset-specific.

%-----------------------------------------------------------------------
\section{Experimental Evaluation}
\label{sec:experimental-evaluation}
%We implemented our method in \meta \cite{gat} and we demonstrate here its effectiveness with an extensive experimental evaluation. 
We consider the main standard datasets used in the fairness literature and we preprocess them by 
following the steps of Ruoss et al.~\cite[Section~5]{ruoss2020learning} for 
their experiments on individual fairness for deep neural networks: (1) standardize numerical attributes to zero mean and unit variance; (2) one-hot encoding of all 
categorical features; (3) drop rows/columns containing missing values; and (4) split into train and test set. These datasets concern binary classification tasks, although our fair learning naturally extends to multiclass classification with no specific effort.  We will make all the code, datasets and preprocessing pipelines of \FATT publicly available upon publication of this work.
\begin{description}
 \item [Adult.] The Adult income dataset \cite{dua2017uci} is extracted from the 1994 US Census database. Every sample assigns a yearly income (below or above \$50K) to an individual based on personal attributes such as gender, race, and occupation.
 \item [Compas.] The COMPAS dataset contains data collected on the use of the COMPAS risk assessment tool in Broward County, Florida \cite{angwin2016machine}. Each sample predicts the risk of recidivism for individuals based on personal attributes and criminal history.
 \item [Crime.] The Communities and Crime dataset \cite{dua2017uci} contains socio-economic, law enforcement, and crime data for communities within the US. Each sample indicates whether a community is above or below the median number of violent crimes per population.
 \item [German.] The German Credit dataset \cite{dua2017uci} contains samples assigning a good or bad credit score to individuals.
 \item [Health.] The heritage Health dataset (\url{https://www.kaggle.com/c/hhp})
 %\cite{health} 
 contains physician records and insurance claims. Each sample predicts the ten-year mortality (above or below the median Charlson index) for a patient.
\end{description}

\begin{table}[ht]
 \resizebox{\columnwidth}{!}{
 \centering
 \begin{tabular}{| l | r | r r | r r |}
  \hline
         &            &\multicolumn{2}{| c |}{\textbf{Training Set}} & \multicolumn{2}{| c |}{\textbf{Test Set}} \\
  \textbf{dataset} & \#features & size & positive & size & positive \\
  \hline
  adult  & 103 &  30162 & 24.9\% & 15060 & 24.6\% \\
  compas & 371 &   4222 & 53.3\% &  1056 & 55.6\% \\
  crime  & 147 &   1595 & 50.0\% &   399 & 49.6\% \\
  german &  56 &    800 & 69.8\% &   200 & 71.0\% \\
  health & 110 & 174732 & 68.1\% & 43683 & 68.0\% \\
  \hline
 \end{tabular}
 }
 \caption{Overview of Datasets.}
 \label{tab:datasets}
\end{table}

Table~\ref{tab:datasets} displays size and distribution of positive samples for these datasets. As noticed by \cite{ruoss2020learning}, some datasets exhibit a highly unbalanced label distribution. For example, for the adult dataset, the constant classifier $C(\vec{x}) = 1$ would achieve $75.4\%$ test set accuracy and $100\%$ individual fairness with respect to any similarity relation. Hence, we follow \cite{ruoss2020learning} and 
we will evaluate and report the balanced accuracy of our \FATT classifiers, i.e., 
\begin{equation*}
\resizebox{\columnwidth}{!}
{%
$0.5\, (\frac{\textit{truePositive}}{\textit{truePositive} + \textit{falseNegative}} + \frac{\textit{trueNegative}}{\textit{trueNegative} + \textit{falsePositive}})$.
}
\end{equation*}

\subsection{Similarity Relations}\label{subsec:similarity}

We consider four different types of similarity relations, as described by Ruoss et al. \cite[Section~5.1]{ruoss2020learning}. 
In the following, let $I \subseteq \mathbb{N}$ denote
the set of indexes of features of an individual after one-hot encoding.
\begin{description}
 \item [\textsc{noise}:] Two individuals $\vec{x}, \vec{y} \in X$ are similar when a subset of their (standardized) numerical features indexed by a given subset $I' \subseteq I$ differs less than a given threshold $\tau \geq 0$, while all the 
other features are unchanged: $(\vec{x}, \vec{y}) \in S_{\textit{noise}}$ iff $| {\vec{x}}_i - {\vec{y}}_i | \leq \tau$ for all $i \in I'$, and ${\vec{x}}_i = {\vec{y}}_i$ for all $i\in I\smallsetminus I'$. For our experiments, we consider $\epsilon = 0.3$ in the standardized input space, e.g., for adult two individuals are similar if their age difference is at most 3.95 years.
 
 \item [\textsc{cat}:] Two individuals are similar if they are identical except for one or more categorical sensitive attributes indexed by $I' \subseteq I$: $(\vec{x}, \vec{y}) \in S_{\textit{cat}}$ iff ${\vec{x}}_i = {\vec{y}}_i$ for all $i \in I\smallsetminus I'$. For adult and german, we select the gender attribute. For compas, we identify race as sensitive attribute. For crime, we consider two individuals similar regardless of their state. Lastly, for health, neither gender nor age group should affect the final prediction.
 
 \item [\textsc{noise-cat}:] Given noise   and categorical  similarity relations $S_{\textit{noise}}$ and $S_{\textit{cat}}$, their union $S_{\textit{noise-cat}} \ud S_{\textit{noise}} \cup S_{\textit{cat}}$ models a relation where two individuals are similar 
when some of their numerical attributes differ up to a given threshold while the other attributes are equal except some categorical features.
 
 \item [\textsc{conditional-attribute}:] Here,  similarity is a disjunction of two mutually exclusive cases. Consider a numerical attribute $\vec{x}_i$, a threshold $\tau\geq 0$ 
and two noise similarities $S_{n_1}, S_{n_2}$. Two individuals are defined to be 
similar if their $i$-th attributes are similar for $S_{n_1}$ and are 
bounded by $\tau$ or these attributes are above $\tau$ and similar 
for $S_{n_2}$: $S_{\textit{cond}} \ud \{(\vec{x}, \vec{y}) \in S_{n_1} ~|~ {\vec{x}}_i \leq \tau,\, {\vec{y}}_i \leq \tau\} \cup \{(\vec{x}, \vec{y}) \in S_{n_2} ~|~ {\vec{x}}_i > \tau,\, {\vec{y}}_i > \tau\}$. For adult, we consider the median age as threshold $\tau = 37$, and two noise similarities based on age with thresholds $0.2$ and $0.4$, which correspond to age differences of $2.63$ and $5.26$ years respectively. For german, we also consider the median age $\tau = 33$ and the same noise similarities on age, that correspond to age differences of $0.24$ and $0.47$ years.
\end{description}

\noindent
Note that our approach is not limited to supporting these similarity relations. Further domain-specific similarities can be defined and handled by our approach by instantiating the underlying verifier \silva with an appropriately over-approximating abstract domain to retain soundness. Moreover, if the similarity relation can be precisely represented in the chosen abstract domain, we also retain completeness.

\subsection{Setup} 
Our experimental evaluation compares CART trees and Random Forests with our \FATT tree 
models. CARTs and RFs are trained by scikit-learn. 
%\cite{scikitlearn}, 
We first run a preliminary phase for tuning the hyper-parameters for CARTs and RFs. In particular, we considered both entropy and Gini index as split criteria, and we 
checked maximum tree depths ranging from $5$ to $100$ with step $10$. For RFs, we scanned the maximum number of trees ($5$ to $100$, step $10$). Cross validation inferred 
the optimal hyper-parameters, where the datasets have been split in $80\%$ training and $20\%$ validation sets. The hyper-parameters of \FATT (i.e, weights of accuracy and fairness in the objective function, type of mutation, selection function, number of iterations)
% eagerness in split evaluation
% \todo{All of this is not clear without saying something more in Section 5}
by assessing convergence speed, maximum fitness value and variance among fitness in the population during the training phase. \FATT trained single decision 
trees rather than forests, thus providing more compact and interpretable models. 
It turned out that accuracy and fairness of single \FATT trees are already competitive, where 
individual fairness may exceed $85\%$ for the most challenging similarities. We therefore concluded that ensembles of \FATT trees do not introduce statistically 
significant benefits over single decision trees.
% \todo{We should probably say why here.}
%
Since \FATT trees are stochastic by relying on random seeds, each experimental 
test has been repeated 1000 times and the results refer to their median value. 

\subsection{Results}

\begin{table}[ht]
 \resizebox{\columnwidth}{!}{
 \begin{tabular}{| l | r r | r r | r r | r r | r r |}
  \hline
         & \multicolumn{2}{| c |}{\multirow{2}{*}{Acc.\ \%}} & \multicolumn{2}{| c |}{\multirow{2}{*}{Bal.Acc.\ \%}} & \multicolumn{6}{| c |}{Individual Fairness $\fairm_{T}$ \%} \\[3pt] \cline{6-11}
         & & & & & \multicolumn{2}{| c |}{\textsc{cat}} & \multicolumn{2}{| c |}{\textsc{noise}} & \multicolumn{2}{| c |}{\textsc{noise-cat}} \\
  \textbf{Dataset} & \multicolumn{1}{c}{RF} & \multicolumn{1}{c|}{\FATT} & \multicolumn{1}{c}{RF} & \multicolumn{1}{c|}{\FATT}  & \multicolumn{1}{c}{RF} & \multicolumn{1}{c|}{\FATT}  & \multicolumn{1}{c}{RF} & \multicolumn{1}{c|}{\FATT}  & \multicolumn{1}{c}{RF} & \multicolumn{1}{c|}{\FATT}  \\
  \hline
  adult            &  82.76 &  80.84 &  70.29 &  61.86 &  91.71 & 100.00 &  85.44 &  95.21 &  77.50 &  95.21 \\
  compas           &  66.57 &  64.11 &  66.24 &  63.83 &  48.01 & 100.00 &  35.51 &  85.98 &  30.87 &  85.98 \\
  crime            &  80.95 &  79.45 &  80.98 &  79.43 &  86.22 & 100.00 &  31.83 &  75.19 &  32.08 &  75.19 \\
  german           &  76.50 &  72.00 &  63.62 &  52.54 &  91.50 & 100.00 &  92.00 &  99.50 &  90.00 &  99.50 \\
  health           &  85.29 &  77.87 &  83.27 &  73.59 &   7.84 &  99.99 &  47.66 &  97.04 &   2.91 &  97.03 \\
  \hline
  \textbf{Average}          &  \textbf{78.41} &  74.85 &  \textbf{72.88} &  66.25 &  65.06 & \textbf{100.00} &  58.49 &  \textbf{90.58} &  46.67 &  \textbf{90.58} \\
  \hline
 \end{tabular}
 }
 \caption{RF and \FATT comparison.}
 \label{tab:comparison}
\end{table}

Table~\ref{tab:comparison} shows a comparison between RFs and {\FATT}s. We show accuracy, balanced accuracy and individual fairness with respect to the \textsc{noise}, \textsc{cat}, and \textsc{noise-cat} similarity relations as computed on the test sets $T$.
As expected, \FATT trees are slightly less accurate than RFs~---~$~3.6\%$ on average, which also reflects to balanced accuracy~---~but outperform them in every fairness test. On average, the fairness increment ranges between $+35\%$ to $+45\%$ among different similarity relations.
Table~\ref{tab:comparison-attribute} shows the comparison 
for the conditional-attribute similarity, which applies to adult and german datasets only. Here, the average fairness increase of \FATT models is $+8.5\%$.

\begin{table}[ht]
\centering
 \resizebox{0.5\columnwidth}{!}{
 \begin{tabular}{| l | c c |}
  \hline
         & \multicolumn{2}{| c |}{Individual Fairness $\fairm_{T}$ \%} \\[4pt]
  \textbf{Dataset} & \multicolumn{1}{c}{RF} & \multicolumn{1}{c|}{\FATT}  \\
  \hline
  adult            &  84.75 &  94.12 \\
  german           &  91.50 &  99.50 \\
  \hline
 \end{tabular}
}
 \caption{Comparison for conditional-attribute.}
 \label{tab:comparison-attribute}
\end{table}

\begin{figure}[ht]
 \centering
 \includegraphics[width=0.5\textwidth]{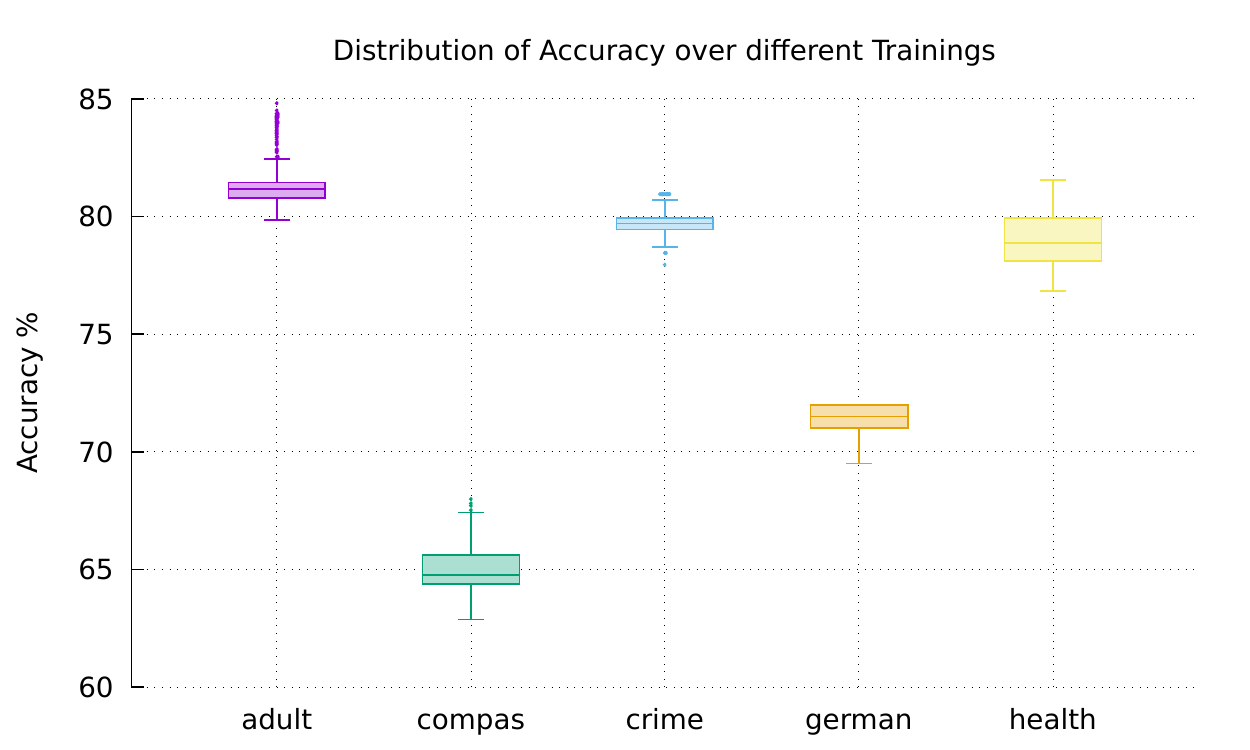}
 \includegraphics[width=0.5\textwidth]{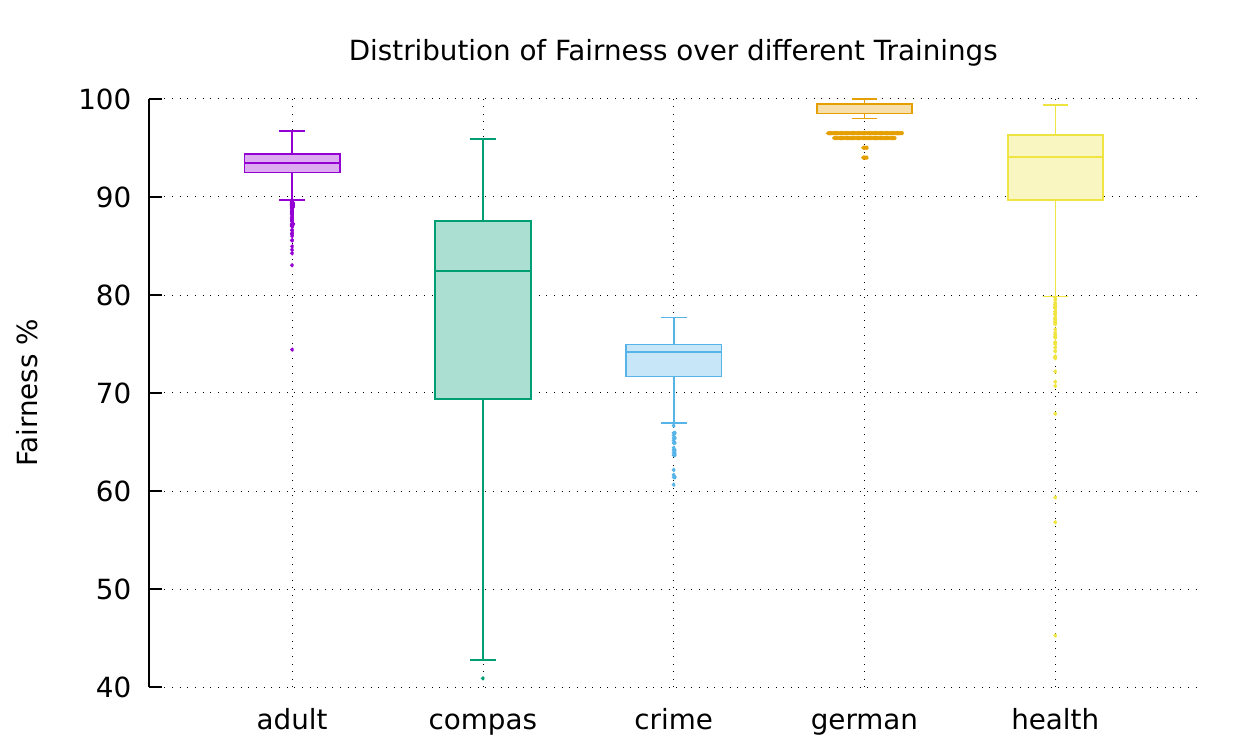}
 \caption{Accuracy (top) and Fairness (bottom).}
 \label{fig:accuracy-fairness}
\end{figure}

Fig.~\ref{fig:accuracy-fairness} shows the distribution of accuracy and individual 
fairness for \FATT trees over 1000 runs of the \FATT  learning algorithm. This plot 
is for fairness with respect to noise-cat similarity, as this is the most 
challenging relation to train for (this is a consequence of~\eqref{individual-fairness-subset}). We can observe a stable behaviour for accuracy, with $\approx50\%$ of the observations laying within one percentile from the median. The results for fairness are analogous, although for compas we report a higher variance of the distribution, where the lowest observed fairness percentage is $\approx10\%$ higher than the corresponding one for RFs. We claim that 
this may depend by the high number of features in the dataset, which makes fair 
training a challenging task.

\begin{table}[ht]
 \centering
 \resizebox{\columnwidth}{!}{
 \begin{tabular}{| l | r r | r r | r r | r r |}
  \hline
         & \multicolumn{2}{| c |}{\multirow{2}{*}{\textbf{Model size}}} & \multicolumn{6}{| c |}{\textbf{Avg.\ verification time per sample (ms)}} \\
         \cline{4-9}
         & & & \multicolumn{2}{| c |}{\textsc{cat}} & \multicolumn{2}{| c |}{\textsc{noise}} & \multicolumn{2}{| c |}{\textsc{noise-cat}} \\
  \textbf{Dataset} &\multicolumn{1}{c}{RF} & \multicolumn{1}{c|}{\FATT} & \multicolumn{1}{c}{RF} & \multicolumn{1}{c|}{\FATT} & \multicolumn{1}{c}{RF} & \multicolumn{1}{c|}{\FATT} & \multicolumn{1}{c}{RF} & \multicolumn{1}{c|}{\FATT}\\
  \hline
  adult            & 1427    & 43 &   0.03 &   0.02 &   0.03 &   0.02 &   0.03 &   0.02 \\
  compas           & 147219  & 75 &   0.36 &   0.07 &   0.47 &   0.07 &   0.61 &   0.07 \\
  crime            & 14148   & 11 &   0.12 &   0.07 & 2025.13 &   0.07 & 2028.47 &   0.07 \\
  german           & 5743    &  2 &   0.06 &   0.03 &   0.06 &   0.02 &   0.07 &   0.03 \\
  health           & 2558676 & 84 &   1.40 &   0.06 &   0.91 &   0.05 &   3.10 &   0.06 \\
  \hline
 \end{tabular}
 }
 \caption{Model sizes and verification times.}
 \label{tab:size-time}
\end{table}

Table~\ref{tab:size-time} compares the size of RF and \FATT models, defined as 
total number of leaves. It turns out that \FATT tree models are 
orders of magnitude smaller and, thus, more interpretable than RFs (while having comparable
accuracy and significantly enhanced fairness). Let us also remark that the average verification time per sample for our \FATT models is always less than $0.01$ milliseconds. 

\begin{table}[ht]
 \centering
 \resizebox{\columnwidth}{!}{
 \begin{tabular}{| l | r r r | r r r | r r r |}
  \hline
          & \multicolumn{3}{| c |}{\textbf{\FATT}} & \multicolumn{3}{| c |}{\textbf{Natural CART}} & \multicolumn{3}{| c |}{\textbf{Hinted CART}} \\
  \textbf{Dataset}  & Acc.\ \% & Fair.\ \% & Size & Acc.\ \% & Fair.\ \% & Size & Acc.\ \% & Fair.\ \% & Size \\
  \hline
  adult    &  80.84 &  95.21 &    43 &  85.32 &  77.56 &   270 &  84.77 &  87.46 &    47 \\
  compas   &  64.11 &  85.98 &    75 &  65.91 &  22.25 &    56 &  65.91 &  22.25 &    56 \\
  crime    &  79.45 &  75.19 &    11 &  77.69 &  24.31 &    48 &  77.44 &  60.65 &     8 \\
  german   &  72.00 &  99.50 &     2 &  75.50 &  57.50 &   115 &  73.50 &  86.00 &     4 \\
  health   &  77.87 &  97.03 &    84 &  83.85 &  79.98 &  2371 &  82.25 &  93.64 &   100 \\
  \hline
  \textbf{Average}  &  74.85 &  \textbf{90.58} &    \textbf{43} &  \textbf{77.65} &  52.32 &   572 &  76.77 &  70.00 &    \textbf{43} \\
  \hline
 \end{tabular}
 }
 \caption{Decision trees comparison.}
 \label{tab:comparison-decision-trees}
\end{table}

Finally, in Table~\ref{tab:comparison-decision-trees} compares \FATT  models with natural CART trees in terms of accuracy, size, and fairness with respect to the noise-cat similarity. While CARTs are approximately $3\%$ more accurate than \FATT on average, they are roughly half less fair and more than ten times larger. 

It is well known that decision trees often overfit \cite{bramer2007} due to their high number of leaves, thus 
yielding unstable/unfair models. 
Post-training techniques such as tree pruning are often used to mitigate overfitting \cite{kearns1998}, although they are deployed when a tree has been already fully trained and thus often pruning is poorly beneficial.  As a byproduct of our approach, we trained a set of natural CART trees, denoted by Hint
in Table~\ref{tab:comparison-decision-trees}, which exploits hyper-parameters as ``hinted''
by \FATT training.  
%rather than following a standard cross-validation phase.  
%where we use hints about the size and shape of a tree as input hyper-parameters for CART learning algorithm. 
In particular, in this ``hinted'' learning of CART trees,
the maximum tree depth and the minimum number of samples per leaf are obtained as 
tree depth and minimum number of samples of our best \FATT models. Interestingly, 
the results in Table~\ref{tab:comparison-decision-trees} show
that 
these ``hinted''  decision trees have roughly the same size of our \FATT trees, are approximately $20\%$ more fair than natural CART trees and just $1\%$ less accurate. 
Overall, it turns out that the  general performance of these ``hinted'' CARTs is halfway between natural CARTs and {\FATT}s, both in term of accuracy and fairness, while having the same compactness of \FATT models.

\section{Conclusion}
\label{sec:conclusion}
We believe that this work contributes to push forward the use of formal verification methods in decision tree learning, in particular a very well known program analysis technique such as abstract interpretation is proved to be successful for training and verifying decision tree classifiers which are both accurate and fair, improve on state-of-the-art CART and random forest models, while being much more compact and interpretable. We also showed how information from our \FATT trees can be exploited to tune the natural training process of decision 
trees. As future work we plan to extend further our fairness analysis by considering alternative fairness definitions, such as group or statistical fairness.

%\newpage
%-----------------------------------------------------------------------
% Bibliography
\bibliography{bibliography}

\begin{thebibliography}{35}
\providecommand{\natexlab}[1]{#1}
\providecommand{\url}[1]{\texttt{#1}}
\providecommand{\urlprefix}{URL }
\expandafter\ifx\csname urlstyle\endcsname\relax
  \providecommand{\doi}[1]{doi:\discretionary{}{}{}#1}\else
  \providecommand{\doi}{doi:\discretionary{}{}{}\begingroup
  \urlstyle{rm}\Url}\fi

\bibitem[{Aghaei, Azizi, and Vayanos(2019)}]{aghaei19}
Aghaei, S.; Azizi, M.~J.; and Vayanos, P. 2019.
\newblock Learning Optimal and Fair Decision Trees for Non-Discriminative
  Decision-Making.
\newblock In \emph{Proceedings of the Thirty-Third {AAAI} Conference on
  Artificial Intelligence, {AAAI} 2019}, 1418--1426. {AAAI} Press.
\newblock \doi{10.1609/aaai.v33i01.33011418}.
\newblock \urlprefix\url{https://doi.org/10.1609/aaai.v33i01.33011418}.

\bibitem[{Andriushchenko and Hein(2019)}]{Andriushchenko19}
Andriushchenko, M.; and Hein, M. 2019.
\newblock {Provably Robust Boosted Decision Stumps and Trees against
  Adversarial Attacks}.
\newblock In \emph{Proc.\ 33rd Annual Conference on Neural Information
  Processing Systems (NeurIPS 2019)}.

\bibitem[{Angwin et~al.(2016)Angwin, Larson, Mattu, and
  Kirchner}]{angwin2016machine}
Angwin, J.; Larson, J.; Mattu, S.; and Kirchner, L. 2016.
\newblock {Machine Bias}.
\newblock \emph{ProPublica, May} 23: 2016.

\bibitem[{Barocas and Selbst(2016)}]{Barocas}
Barocas, S.; and Selbst, A.~D. 2016.
\newblock {Big Data's Disparate Impact}.
\newblock \emph{California Law Review} 104: 671.

\bibitem[{Bertsimas and Dunn(2017)}]{Bertsimas17}
Bertsimas, D.; and Dunn, J. 2017.
\newblock Optimal classification trees.
\newblock \emph{Mach. Learn.} 106(7): 1039--1082.
\newblock
  \urlprefix\url{http://dblp.uni-trier.de/db/journals/ml/ml106.html#BertsimasD17}.

\bibitem[{Bramer(2007)}]{bramer2007}
Bramer, M. 2007.
\newblock Avoiding overfitting of decision trees.
\newblock \emph{Principles of data mining} 119--134.

\bibitem[{Breiman(2001)}]{breiman-random-forests}
Breiman, L. 2001.
\newblock {Random Forests}.
\newblock \emph{Machine Learning} 45(1): 5--32.
\newblock \doi{10.1023/A:1010933404324}.
\newblock \urlprefix\url{https://doi.org/10.1023/A:1010933404324}.

\bibitem[{Breiman et~al.(1984)Breiman, Friedman, Olshen, and
  Stone}]{BreimanFOS84}
Breiman, L.; Friedman, J.~H.; Olshen, R.~A.; and Stone, C.~J. 1984.
\newblock \emph{{Classification and Regression Trees}}.
\newblock Wadsworth.
\newblock ISBN 0-534-98053-8.

\bibitem[{Calzavara, Lucchese, and Tolomei(2019)}]{calzavara2019B}
Calzavara, S.; Lucchese, C.; and Tolomei, G. 2019.
\newblock {Adversarial Training of Gradient-Boosted Decision Trees}.
\newblock In \emph{Proc.\ 28th ACM International Conference on Information and
  Knowledge Management (CIKM 2019)}, 2429--2432.
\newblock ISBN 978-1-4503-6976-3.
\newblock \doi{10.1145/3357384.3358149}.
\newblock \urlprefix\url{http://doi.acm.org/10.1145/3357384.3358149}.

\bibitem[{Calzavara et~al.(2020)Calzavara, Lucchese, Tolomei, Abebe, and
  Orlando}]{calzavara20}
Calzavara, S.; Lucchese, C.; Tolomei, G.; Abebe, S.~A.; and Orlando, S. 2020.
\newblock {T}{R}{E}{A}{N}{T}: training evasion-aware decision trees.
\newblock \emph{Data Mining and Knowledge Discovery}
  \doi{10.1007/s10618-020-00694-9}.
\newblock \urlprefix\url{https://doi.org/10.1007/s10618-020-00694-9}.

\bibitem[{Carlini and Wagner(2017)}]{carlini}
Carlini, N.; and Wagner, D.~A. 2017.
\newblock {Towards Evaluating the Robustness of Neural Networks}.
\newblock In \emph{Proc.\ of 38th {IEEE} Symposium on Security and Privacy (S
  \& P 2017)}, 39--57.
\newblock \doi{10.1109/SP.2017.49}.
\newblock \urlprefix\url{https://doi.org/10.1109/SP.2017.49}.

\bibitem[{Chen et~al.(2019)Chen, Zhang, Boning, and Hsieh}]{ChenZBH19}
Chen, H.; Zhang, H.; Boning, D.~S.; and Hsieh, C. 2019.
\newblock Robust Decision Trees Against Adversarial Examples.
\newblock In \emph{Proc.\ 36th Int.\ Conf.\ on Machine Learning, (ICML 2019)},
  1122--1131.
\newblock \urlprefix\url{http://proceedings.mlr.press/v97/chen19m.html}.

\bibitem[{Chouldechova(2017)}]{Chouldechova}
Chouldechova, A. 2017.
\newblock {Fair Prediction with Disparate Impact: A Study of Bias in Recidivism
  Prediction Instruments}.
\newblock \emph{Big Data} 5(2): 153--163.
\newblock \doi{10.1089/big.2016.0047}.
\newblock \urlprefix\url{https://doi.org/10.1089/big.2016.0047}.

\bibitem[{Cousot and Cousot(1977)}]{CC77}
Cousot, P.; and Cousot, R. 1977.
\newblock Abstract interpretation: a unified lattice model for static analysis
  of programs by construction or approximation of fixpoints.
\newblock In \emph{Proc.\ 4th ACM Symposium on Principles of Programming
  Languages (POPL 1977)}, 238--252.
\newblock \doi{10.1145/512950.512973}.
\newblock \urlprefix\url{http://doi.acm.org/10.1145/512950.512973}.

\bibitem[{Dua and Graff(2017)}]{dua2017uci}
Dua, D.; and Graff, C. 2017.
\newblock UCI machine learning repository.

\bibitem[{Dwork et~al.(2012)Dwork, Hardt, Pitassi, Reingold, and
  Zemel}]{dwork2012fairness}
Dwork, C.; Hardt, M.; Pitassi, T.; Reingold, O.; and Zemel, R. 2012.
\newblock {Fairness Through Awareness}.
\newblock In \emph{Proc.\ 3rd Innovations in Theoretical Computer Science
  Conference}, 214--226.

\bibitem[{Friedman(2001)}]{friedman2001greedy}
Friedman, J.~H. 2001.
\newblock {Greedy Function Approximation: A Gradient Boosting Machine}.
\newblock \emph{Annals of statistics} 1189--1232.

\bibitem[{Goodfellow, McDaniel, and Papernot(2018)}]{cacm18}
Goodfellow, I.; McDaniel, P.; and Papernot, N. 2018.
\newblock {Making Machine Learning Robust Against Adversarial Inputs}.
\newblock \emph{Commun. ACM} 61(7): 56--66.
\newblock ISSN 0001-0782.
\newblock \doi{10.1145/3134599}.
\newblock \urlprefix\url{http://doi.acm.org/10.1145/3134599}.

\bibitem[{Grari et~al.(2020)Grari, Ruf, Lamprier, and Detyniecki}]{Grari}
Grari, V.; Ruf, B.; Lamprier, S.; and Detyniecki, M. 2020.
\newblock {Achieving Fairness with Decision Trees: An Adversarial Approach}.
\newblock \emph{Data Sci. Eng.} 5(2): 99--110.
\newblock \doi{10.1007/s41019-020-00124-2}.
\newblock \urlprefix\url{https://doi.org/10.1007/s41019-020-00124-2}.

\bibitem[{Hardt, Price, and Srebro(2016)}]{Hardt}
Hardt, M.; Price, E.; and Srebro, N. 2016.
\newblock Equality of Opportunity in Supervised Learning.
\newblock In \emph{Proc.\ 30th Annual Conference on Neural Information
  Processing Systems (NeurIPS 2016)}, 3315--3323.
\newblock
  \urlprefix\url{http://papers.nips.cc/paper/6374-equality-of-opportunity-in-supervised-learning}.

\bibitem[{Holland(1984)}]{holland1984genetic}
Holland, J.~H. 1984.
\newblock Genetic algorithms and adaptation.
\newblock In \emph{Adaptive Control of Ill-Defined Systems}, 317--333.
  Springer.

\bibitem[{Kantchelian, Tygar, and Joseph(2016)}]{kantchelian}
Kantchelian, A.; Tygar, J.~D.; and Joseph, A.~D. 2016.
\newblock {Evasion and Hardening of Tree Ensemble Classifiers}.
\newblock In \emph{Proc.\ 33rd International Conference on Machine Learning
  (ICML 2016)}, 2387--2396.
\newblock \urlprefix\url{http://dl.acm.org/citation.cfm?id=3045390.3045642}.

\bibitem[{Kearns and Mansour(1998)}]{kearns1998}
Kearns, M.~J.; and Mansour, Y. 1998.
\newblock A Fast, Bottom-Up Decision Tree Pruning Algorithm with Near-Optimal
  Generalization.
\newblock In \emph{Proceedings of the Fifteenth International Conference on
  Machine Learning (ICML 1998)}, 269--277.

\bibitem[{Khandani, Kim, and Lo(2010)}]{Khandani}
Khandani, A.~E.; Kim, A.~J.; and Lo, A.~W. 2010.
\newblock {Consumer Credit-Risk Models via Machine-Learning Algorithms}.
\newblock \emph{Journal of Banking \& Finance} 34(11): 2767--2787.
\newblock \doi{https://doi.org/10.1016/j.jbankfin.2010.06.001}.

\bibitem[{Raff, Sylvester, and Mills(2018)}]{Raff}
Raff, E.; Sylvester, J.; and Mills, S. 2018.
\newblock Fair Forests: Regularized Tree Induction to Minimize Model Bias.
\newblock In \emph{Proc. 1st AAAI/ACM Conference on AI, Ethics, and Society
  (AIES 2018)}, 243–250.
\newblock \doi{10.1145/3278721.3278742}.
\newblock \urlprefix\url{https://doi.org/10.1145/3278721.3278742}.

\bibitem[{Ranzato and Zanella(2020{\natexlab{a}})}]{DBLP:conf/aaai/RanzatoZ20}
Ranzato, F.; and Zanella, M. 2020{\natexlab{a}}.
\newblock Abstract Interpretation of Decision Tree Ensemble Classifiers.
\newblock In \emph{Proc.\ 34th {AAAI} Conference on Artificial Intelligence
  (AAAI 2020), {\rm Github:}
  \url{https://github.com/abstract-machine-learning/silva}}, 5478--5486.
\newblock
  \urlprefix\url{https://aaai.org/ojs/index.php/AAAI/article/view/5998}.

\bibitem[{Ranzato and Zanella(2020{\natexlab{b}})}]{gat}
Ranzato, F.; and Zanella, M. 2020{\natexlab{b}}.
\newblock Genetic Adversarial Training of Decision Trees.
\newblock \emph{arXiv:2012.11352, {\rm Github:}
  \url{https://github.com/abstract-machine-learning/meta-silvae}} .

\bibitem[{Rival and Yi(2020)}]{rival-yi}
Rival, X.; and Yi, K. 2020.
\newblock \emph{Introduction to Static Analysis: An Abstract Interpretation
  Perspective}.
\newblock The {M}{I}{T} {P}ress.

\bibitem[{Roh et~al.(2020)Roh, Lee, Whang, and Suh}]{roh20}
Roh, Y.; Lee, K.; Whang, S.; and Suh, C. 2020.
\newblock FR-Train: {A} Mutual Information-Based Approach to Fair and Robust
  Training.
\newblock In \emph{Proceedings of the 37th International Conference on Machine
  Learning, {ICML} 2020, 13-18 July 2020, Virtual Event}, volume 119 of
  \emph{Proceedings of Machine Learning Research}, 8147--8157. {PMLR}.
\newblock \urlprefix\url{http://proceedings.mlr.press/v119/roh20a.html}.

\bibitem[{Ruoss et~al.(2020)Ruoss, Balunovic, Fischer, and
  Vechev}]{ruoss2020learning}
Ruoss, A.; Balunovic, M.; Fischer, M.; and Vechev, M. 2020.
\newblock Learning Certified Individually Fair Representations.
\newblock In \emph{Proc.\ 34th Annual Conference on Advances in Neural
  Information Processing Systems (NeurIPS 2020)}.

\bibitem[{Schumann et~al.(2020)Schumann, Foster, Mattei, and
  Dickerson}]{Schumann}
Schumann, C.; Foster, J.~S.; Mattei, N.; and Dickerson, J.~P. 2020.
\newblock {We Need Fairness and Explainability in Algorithmic Hiring}.
\newblock In \emph{Proc.\ 19th International Conference on Autonomous Agents
  and Multiagent Systems (AAMAS 2020)}, 1716--1720.
\newblock \urlprefix\url{https://dl.acm.org/doi/abs/10.5555/3398761.3398960}.

\bibitem[{{Srinivas} and {Patnaik}(1994)}]{Srinivas}
{Srinivas}, M.; and {Patnaik}, L.~M. 1994.
\newblock Genetic algorithms: a survey.
\newblock \emph{Computer} 27(6): 17--26.

\bibitem[{Urban et~al.(2020)Urban, Christakis, W{\"{u}}stholz, and
  Zhang}]{Urban}
Urban, C.; Christakis, M.; W{\"{u}}stholz, V.; and Zhang, F. 2020.
\newblock {Perfectly Parallel Fairness Certification of Neural Networks}.
\newblock \emph{Proceedings of the ACM on Programming Languages} 4({OOPSLA}):
  185:1--185:30.

\bibitem[{Yurochkin, Bower, and Sun(2020)}]{yurochkin20}
Yurochkin, M.; Bower, A.; and Sun, Y. 2020.
\newblock Training individually fair {ML} models with sensitive subspace
  robustness.
\newblock In \emph{8th International Conference on Learning Representations,
  {ICLR} 2020, Addis Ababa, Ethiopia, April 26-30, 2020}. OpenReview.net.
\newblock \urlprefix\url{https://openreview.net/forum?id=B1gdkxHFDH}.

\bibitem[{Zafar et~al.(2017)Zafar, Valera, Gomez{-}Rodriguez, and
  Gummadi}]{Zafar}
Zafar, M.~B.; Valera, I.; Gomez{-}Rodriguez, M.; and Gummadi, K.~P. 2017.
\newblock Fairness Beyond Disparate Treatment {\&} Disparate Impact: Learning
  Classification without Disparate Mistreatment.
\newblock In \emph{Proc.\ 26th International Conference on World Wide Web (WWW
  2017)}, 1171--1180.
\newblock \doi{10.1145/3038912.3052660}.
\newblock \urlprefix\url{https://doi.org/10.1145/3038912.3052660}.

\end{thebibliography}

\end{document}